\documentclass[11pt,a4paper]{article}
\usepackage[hyperref]{emnlp2020}
\usepackage{times}
\usepackage{latexsym}

\usepackage[utf8]{inputenc}
\usepackage{microtype}

\aclfinalcopy 

\usepackage{amsmath}
\usepackage{booktabs} 
\usepackage{graphicx} 
\usepackage{subfig} 
\usepackage{multirow} 
\usepackage{latexsym}

\usepackage{xcolor,soul} 

\DeclareRobustCommand{\hlcyan}[1]{{\sethlcolor{cyan}\hl{#1}}}
\DeclareRobustCommand{\hlpink}[1]{{\sethlcolor{pink}\hl{#1}}}
\DeclareRobustCommand{\hlgray}[1]{{\sethlcolor{lightgray}\hl{#1}}}

\newcommand{\entlabel}{\textsc{\hlcyan{ent}}}
\newcommand{\neulabel}{\textsc{\hlgray{neu}}}
\newcommand{\conlabel}{\textsc{\hlpink{con}}}

\newcommand{\emp}{$\rule{1em}{0.15mm}$~}

\title{The Extraordinary Failure of Complement Coercion Crowdsourcing}

\author{Yanai Elazar \, Victoria Basmov \, Shauli Ravfogel \, Yoav Goldberg \, Reut Tsarfaty \\
Computer Science Department, Bar Ilan University \\
Allen Institute for Artificial Intelligence \\
  {\tt  \{yanaiela,vikasaeta,shauli.ravfogel\}@gmail.com}\\
  {\tt \{yoav.goldberg,reut.tsarfaty\}@gmail.com}}

\date{}

\begin{document}

\maketitle
\begin{abstract}
Crowdsourcing has eased and scaled up the collection of linguistic annotation in recent years. In this work, we follow known methodologies of collecting labeled data for the \textit{complement coercion} phenomenon.
These are constructions with an \emph{implied} action --- e.g., ``I started a new book I bought last week'', where the implied action is \emph{reading}. We aim to collect annotated data for this phenomenon by reducing it to either of two known tasks:  Explicit Completion and Natural Language Inference.
However, in both cases, crowdsourcing resulted in low agreement scores, even though we followed the same methodologies as in previous work. Why does the same process fail to yield high agreement scores?
We specify our modeling schemes, highlight the differences with previous work and provide some  insights about the task and possible explanations for the failure.
We conclude that specific phenomena require tailored solutions, not only in specialized algorithms, but also in data collection methods.
\end{abstract}

\section{Introduction}

\begin{table}[t!]
\centering
\resizebox{\columnwidth}{!}{%

\begin{tabular}{ll}
\toprule
 Task &  Annotations \\
\midrule

\underline{Explicit} & \\

After a heartfelt vow, she agrees & \multirow{3}{*}{$\{$officiating$\}, \phi$} \\
and the two begin kissing as the preacher & \\ 
tries to \textbf{continue} \emp{} the ceremony. & \\

\midrule

\underline{Entailment} & \\

Hunter waited for max to \textbf{finish}\\ his burger before asking him again. $\leadsto$ &  \multirow{2}{*}{\entlabel{} \neulabel{} \conlabel{}} \\ 
Hunter waited for max to \textbf{finish swallowing}\\ his burger before asking him again. &\\

\bottomrule
\end{tabular}
}
\caption{Examples for the two modeling and annotation schemes used in this work. Both examples are labeled with different (disagreeing) answers. 
In the Explicit modeling, each label is a set, which can be empty ($\phi$) (meaning that no event is implied), or not (and thus the context suggests an implied event).
The second modeling follows the NLI scheme, a standard approach for evaluating language understanding.
The \entlabel{}, \neulabel{} and \conlabel{} labels refer to the entail, neutral and contradict labels accordingly.}

\label{tbl:example}

\end{table}

Crowdsourcing has become extremely popular in recent years for annotating datasets.
Many works use frameworks like Amazon Mechanical Turk (AMT) by converting complex linguistic tasks into easy-to-grasp presentations which make it possible to crowdsource linguistically-annotated data at scale \cite{snli:emnlp2015,qasrl,dasigi2019quoref,Wolfson2020Break}.

In this work, we attempt to use existing methodologies for crowdsourcing linguistic annotations in order to collect annotations for \textit{complement coercion} \cite{Pustejovsky1991TheGL,pustejovsky1995gl}, a phenomenon involving an implied action triggered by an event-selecting verb. 
Specifically, certain \textbf{verb classes} require an \textit{event-denoting} complement, as in: ``I {\bf started} {\em reading} a book'', ``I {\bf finished} {\em eating} the cake'', etc. However, such {\em event-denoting} complements might remain implicit, not appearing in the surface form. Consider for instance, the sentence ``I \textbf{started} \emp a new book.''
Here the event that was started remains implicit.
Our task is then, first, to detect that the verb `{\bf started}' in this  context implies some unmentioned event, and that probable events in this context are \textit{reading} or \textit{writing}. Furthermore, we wish to predict that for ``I \textbf{started} \emp the book I bought yesterday'', the more probable event is \textit{reading}, rather than \textit{writing}.

This phenomenon (described in detail in Section \ref{sec:coercion}) seems intuitive at first, and easy-to-grasp by non-experts. 
However, we find that collecting annotated data for this task via crowdsourcing is very challenging, achieving low agreement scores between annotators (\S \ref{sec:nli}), despite using two common collection methods in frequently used setups. 
The two framings we use for data collection along with examples for them are presented in Table \ref{tbl:example}.

These low agreement scores come as a surprise, given the large body of previous work on crowdsourcing linguistic annotations. 
Why do such issues arise when  collecting data for  {\em complement coercion}, while for similar phenomena the same approaches yield successful results? Although it is difficult to answer this question, we aim to highlight the similarities and the differences with other tasks, and provide some insights into this question.

\section{Background}
\label{sec:coercion}

\paragraph{Complement Coercion}

We are interested in the linguistic phenomenon of {\em complement coercion}.\footnote{Complement coercion has been studied in linguistics from many theoretical viewpoints. See Appendix \ref{sec:app_ling} for background.} 
In complement coercion, there is a clash between an  expectation for a verb argument denoting an event, and the appearance of a noun argument denoting an entity.
Uncovering the covert event requires the comprehender to infer the implied event by invoking the comprehender's lexical semantics and/or world knowledge \cite{Zarcone}.

Consider Examples \ref{ex1} and \ref{ex2}  below, with an implicit event of \textit{reading} or \textit{writing} missing in the surface form.  Inferring the implicit event (marked \emp) is necessary in order to construe the full semantics of this sentence.
\begin{enumerate}
    \item \label{ex1} I \textbf{started} \emp a new book.
    \item \label{ex2} I \textbf{started} \emp a new book I bought last week.
\end{enumerate}

The reconstruction of the covert event requires an interplay between semantics\footnote{E.g., understanding the difference between entity-denoting and event-denoting elements.} and world knowledge.	
In  example \ref{ex1} above, the prefix ``I started \emp'' with the event-selecting verb \textit{started} triggers expectations for some event-denoting object (\textit{reading}, \textit{writing}, \textit{eating}, \textit{watching}, etc). The object that follows, ``a new book'',  narrows down the expectations --- based on world knowledge. As \newcite{mcgregor-etal-2017-geometric} puts it, ``Different nouns grant privileged access to different activities, particularly those which are most frequently performed with the entities they denote''.
Although the entity narrows down the set of possible events, the implied event might remain ambiguous (in Example \ref{ex1}, both \textit{reading} and \textit{writing} are plausible, but \textit{eating} is not). As can be seen in  Example \ref{ex2}, additional context, as in  ``I bought last week'', provides further world-knowledge cues, towards accessing a more specific event (in this case \textit{reading} is more likely than \textit{writing}), thus resolving the remaining ambiguity.

Complement coercion is particularly frequent with certain verb classes, including {\em aspectual verbs} --- verbs that ``describe the initiation, termination, or continuation of an activity" \cite{levin_1993} --- such as: `start', `begin', `continue' and `finish' \cite{mcgregor-etal-2017-geometric}. This set of verbs is the focus of our work.
Note however, that such verbs may appear in similar constructions   that do {\em not} imply any covert action or event. 
For instance, in the following sentence:
\begin{enumerate}
\setcounter{enumi}{2}
    \item \label{ex3} I \textbf{started} a new company.
\end{enumerate}
Here, the verb `start' is used as an entity-selecting (and not event-selecting) verb, a synonym of `found' or `establish'. See more examples of similar non-coercive constructions in Appendix \ref{sec:app_neg}.

Annotated data for complement coercion \cite{pustejovssemevalky2010} was collected in the past, based on a tailor-made annotation methodology \cite{pustejovsky2009glml}, consisting of a multi-step process that includes word-sense disambiguation by experts.
The annotation focused on coercion detection (as well as labeling the arguments type) and did not involve identifying the implied action. 
Here, we aim to collect  complement coercion data via non-expert annotation, at scale, to test whether models can recover the implicit events and resolve the emerging ambiguities.

\paragraph{Crowdsourcing NLI}

NLI, originally framed as Recognizing Textual Entailment (RTE), has become a standard framework for testing reasoning capabilities of models. It originated from the work by \citet{dagan2005pascal}, where a small dataset was curated by experts using precise guidelines with a specific focus on lexical and syntactic variability rather than delicate logical issues, while dismissing cases of disagreements or ambiguity.
\citet{snli:emnlp2015,mnli} then scaled up the task and crowdsourced large-scale NLI datasets. In contrast to \citet{dagan2005pascal}, the task definitions were short and loose, relying on the annotators' common sense understanding. Many works since have been using the NLI framework and the crowdsourcing procedure associated with it to test models for different language phenomena \cite{marelli2014sick,lai2017natural,naik2018stress,ross2019well,yanaka-monotonicity}.

\section{Copmlement Coercion Crowdsourcing}
\label{sec:nli}

\subsection{Explicit Completion Attempt}

We begin by directly modeling  the phenomenon. For a set of sentences containing possibly-coercive verbs, we wish to determine for each  verb if it entails an implicit event, and if so, to figure out what the event is. This direct task-definition approach is reminiscent of studies that collected annotated data for other missing elements phenomena, such as Verb-Phrase Ellipsis \cite{bos2011ellipsis}, Numeric Fused-Heads \cite{nfh}, Bridging \cite{BASHI_bridging,Hou_thesis_bridging} and Sluicing \cite{hansen2020sluicing}. However, when attempting to crowdsource and label complement coercion instances, we reach very low agreement scores in the first step: determining whether there is an implied event or not. We discuss this experiment in greater detail in Appendix \ref{sec:app:explicit}.

\subsection{NLI for Complement Coercion}
\label{sec:nli-cc}

In light of the low agreements on explicit modeling of the task of complement coercion, we turn to a different crowdsourcing approach which was proven successful for many linguistic phenomena -- using NLI as discussed above (\S\ref{sec:coercion}). 
NLI was used to collect data for a wide range of linguistic phenomena: Paraphrase Inference, Anaphora Resolution, Numerical Reasoning, Implicatures and more \cite{white2017inference,poliak2018collecting,jeretic-etal-2020-natural,yanaka-monotonicity,naik2018stress} (see \citet{Poliak2020ASO}).
Therefore, we take a similar approach, with similar methodologies, and make use of NLI as an evaluation setup for the complement coercion phenomenon.

Here we do not directly model the identification and recovery of event verbs, but rather, we  reduce it to an NLI task. Intuitively, if in Example \ref{ex2} the semantically plausible implied event is \textit{reading}, we expect the sentence ``I \textbf{started} a book I bought last week'' to \emph{entail} a sentence that contains the event explicitly: ``I \textbf{started} \emph{reading} a book I bought last week" (Table \ref{tbl:nli}).\footnote{We follow \citet{snli:emnlp2015}, who modeled entailment based on event coreference.} In contrast, we expect ``I \textbf{started} a book'' to be \textit{neutral} with respect to ``I \textbf{started} \emph{reading} a book", since both \textit{reading} and \textit{writing} are plausible in that context, and there is no reason to prefer one of these complements over the other.  
Examples of this format, along with the different labels we employ, are shown in Table \ref{tbl:nli}.

\paragraph{Corpus Candidates}
In order to keep the task simple, we avoid complexities of lexical, semantic and grammatical differences. Each example is composed of a minimal-pair \cite{kaushik2019learning,warstadt2020blimp,Gardner2020contrastSets} consisting of two sentences; one as the premise and the other as the hypothesis. We construct minimal pairs as follows:
First, we extract dependency-parsed sentences from the Book Corpus \cite{book-corpus} containing the lemma of one of the verbs: `start', `begin', `continue' and `finish'.\footnote{These are frequent verbs that often appear in complement coercion constructions \cite{mcgregor-etal-2017-geometric}.} Then, we keep sentences where the anchor verb is attached to another verb with an `xcomp' dependency\footnote{We use spaCy's parser \cite{honnibal2015improved,honnibal2017spacy}.} (e.g. `started' in ``started reading''). These sentences are used as the hypotheses.
To construct the premises, we remove the dependent verb (e.g. `read'), as well as all the words between the anchor and the dependent verb (e.g. `to' in the infinitive form: ``to read''). Additional examples are provided in Appendix \ref{sec:nli-additional}.

Note that this procedure sometimes generates ungrammatical or implausible sentences, which are flagged by the annotators. 

\begin{table}[t!]
\centering
\resizebox{\columnwidth}{!}{%

\begin{tabular}{ll}
\toprule
 Example &  Label \\
\midrule

I started a book I bought last week. $\leadsto$ & \\

I started reading a book I bought last week. & \entlabel{} \\

\midrule

I started a book. $\leadsto$ & \\

I started reading a book. & \neulabel \\ 
I started eating a book. & \conlabel \\

\bottomrule
\end{tabular}
}
\caption{Examples for NLI pairs with a complement coercion structure.
The \entlabel{}, \neulabel{} and \conlabel{} labels refers to entail, neutral and contradict accordingly.}
\label{tbl:nli}

\end{table}

\paragraph{Crowdsourcing Procedure} We follow the standard procedure of collecting NLI data with crowdsourcing and collect annotations from Amazon Mechanical Turk (AMT). Specifically, we follow the instruction from \citet{glockner-vered-nli}, which involves three questions:
\begin{enumerate}
    \item Do the sentences describe the same event?
    \item Does the new sentence add new
information to the original sentence?
    \item Is the new sentence incorrect/ungrammatical?
\end{enumerate}
We discard any example which at least one worker marked as incorrect/ungrammatical. If the answer to the first question was negative, we considered the label as contradict.
Otherwise, we considered the label as entail
if the answer to the second question was negative,
and neutral if it was positive.
A screenshot of the interface is displayed in Figure \ref{fig:nli} in the Appendix. 

We require an approved rate of at least 99\%, at least 5000 completed HITs, and filter workers to be from English-speaking countries. We also condition the turkers to pass a validation test with a perfect score. We pay 8 cents per HIT.

\paragraph{Results}
We collect 76\footnote{We stopped at 76 examples since we did not see fit to annotate more data with the low agreements we obtained.} pairs (after filtering ungrammatical sentences), each labeled by three different annotators. The Fleiss Kappa \cite{fleiss_kappa} agreement is $k=0.24$.
This score is remarkably low, compared to previous work that similarly collected NLI labels and achieved scores between $0.61$ and $0.7$.
Why does this happen? Consider the following examples, along with their labels:

\begin{enumerate}
\setcounter{enumi}{3}
    
    \item \label{nli-easy-ex1} ``We \textbf{finished} Letterman and I got up from the couch and said, I'm going to bed.'' $\leadsto$\\
    ``We \textbf{finished} \textit{watching} Letterman and I got up from the couch and said, I'm going to bed.''\\
    \entlabel{}~\entlabel{}~\entlabel{}

    \item \label{nli-hard-ex1} ``Flo set the sack of sausage and egg biscuits on the counter right as the young man \textbf{finished} his case.'' $\leadsto$\\
    ``Flo set the sack of sausage and egg biscuits on the counter right as the young man \textbf{finished} \textit{pleading} his case.''\\
    \entlabel{}~\neulabel{}~\conlabel{}    
    \item \label{nli-hard-ex2} ``We \textbf{start} the interviews later today.'' $\leadsto$\\
    ``We \textbf{start} \textit{shooting} the interviews later today.''\\
    \neulabel{}~\conlabel{}~\conlabel{}
\end{enumerate}

Example \ref{nli-easy-ex1} was labeled by all three annotators as \textit{entail}.
However, annotators were in disagreement on examples \ref{nli-hard-ex1}, \ref{nli-hard-ex2}.
Example \ref{nli-hard-ex1} was annotated with all three possible labels (entail, contradict and neutral). Indeed, different readings of this phrase are possible --- more formally, different readers  \textit{construe}  the meaning of the utterance differently; ``\textit{[Construal] is a dynamic process of meaning construction, in which speakers and hearers encode and decode, respectively}'' \cite{trott2020construing}.
An annotator who understands the word `case' as a legal case, will choose \emph{entail}, while an annotator who interprets `case' as a bag and imagines a different background story
(for example, a young man packing a brief-case), will choose \emph{contradict}. Finally, an annotator who thinks of both scenarios will choose \emph{neutral}, which can be argued to be the correct answer. However, we find that for a human hearer, holding  both scenarios in mind at the same time is hard, which we attribute to
the  \textit{construal} of meanings. When a human construes an interpretation, they construes it in a single fashion  until primed otherwise. So, it is not natural to conceive competing meaning scenarios when one is already ``locked in'' on a specific construal.

Although the sentence pairs were carefully built to exclude lexical and syntactic variances, ambiguous sentences such as the above recur throughout the dataset.
We believe that these disagreements are inherent to this type of problem, and are not due to other factors such as poor annotations. 
As evidence, the authors of this work also annotated a subset of these examples and reached a similar (low) agreement.

\section{Discussion}
\paragraph{Inherent Disagreements in Human Textual Inferences}
Recently, \citet{pavlick2019inherent} discussed a similar trend of disagreements in five popular NLI datasets (RTE \cite{dagan2005pascal}, SNLI \cite{snli:emnlp2015}, MNLI \cite{mnli}, JOCI \cite{zhang2017ordinal} and DNC \cite{poliak2018collecting}). In their study, annotators had to select the degree to which a premise entails a hypothesis, on a scale \cite{uncertain-nli} (instead of discrete labels).
\citet{pavlick2019inherent} show that even though these datasets are reported to have high agreement scores, specific examples suffer from inherent disagreements. For instance, in about 20\% of the inspected examples, ``there is a nontrivial second component'' (e.g. entailment and neutral).
Our findings are related to theirs, although not identical: while the disagreements they report are due to the individuals' interpretations of a situation, in our case, disagreements are due to the difficulty in imagining a different scenario.
While some works propose to collect annotator disagreements and use them as inputs \cite{plank2014learning,palomaki2018case} (see \citet{pavlick2019inherent} for an elaborated overview), this will not hold in our case, because only one of the labels is typically correct.

However, the bottom-line is the same: these disagreements cannot be dismissed as `noise', they are more profound.
We hypothesize that when tackling specific phenomena like the one we address in this work, which involve sources of disagreements that are often `ignored' (not intentionally) during the collection of large datasets,\footnote{Due to large scale annotations, `marginal' phenomena might be ignored to keep the instructions clear and concise.} these sources of disagreements are highlighted and manifest themselves more clearly. This results in low agreement scores as we see in our study.\\
\textbf{Scale Annotations}\,
Recent works have proposed to collect labels for NLI pairs on a scale \cite{pavlick2019inherent,uncertain-nli,multi-answers-nli}. Although we agree that this technique may produce a more fine-grained understanding of human judgments, \citet{pavlick2019inherent,multi-answers-nli} observed that scale annotations may result in a multi-modality of the distribution. The different distributions can be viewed as different construals, where each individual interprets the example differently. 
\\\textbf{Task Definition}\,
Another issue might arise from the task definition itself. As opposed to annotation efforts for linguistic tasks such as parsing \cite{penn-treebank} and semantic role labeling \cite{carreras2005introduction} that are carried out by expert annotators and often have annotation guidelines of dozens of pages, the transition to crowdsourcing has reduced the guidelines to a few phrases, and expert annotators have been replaced by laymen. This transition required to simplify the guidelines and to avoid complex definition and corner-cases.
Even though crowdsourcing enabled an easier annotation process and collection of huge amounts of data, it also came with a cost: lack of refined definitions and relying on people's ``common sense'' and ``intuition''. However, as we see in this work, such intuitions are not consistent across individuals and are not sufficient for some tasks. We believe that, similar to the issues mentioned above, the lack of proper definitions tends to amplify disagreements when dealing with specific phenomena, which was often the reason behind the elaborated and long guidelines in classic datasets \cite{kalouli2019explaining}.\\
\textbf{Possible Solution}\,
As we approach ``solving'' current NLP dataset, which were once perceived as complicated, we also reach an understanding that the datasets at hand do not reflect the full capacity of language, and specific linguistic phenomena, which may posses specific challenges, are lost in the crowds. Some phenomena turn out to be more complex, and require specific solutions. In this work we show that, like we do with algorithmic solutions 
we need to reconsider the data collection process. We hold that data collection for these phenomena also require training of the annotators \cite{roit2020controlled,pyatkin_qadiscourse}, whether experts or crowdsourcing workers, and may also require coming up with novel annotation protocols.

Another potential solution is to use deliberation between the workers as a mean to improve agreement \cite{schaekermann2018resolvable}. With respect to the disagreements we observed, a deliberation between workers would allow them to share the construals each individual had imagined, thus reaching a consensus on the labels. It would also serve as a training for recovering more construals, allowing them to better identify the \textit{neutral} cases.

\vspace{-0.2em}
\section{Conclusions}
\vspace{-0.2em}

In this work, we attempt to crowdsource annotations for complement coercion constructions. We use two modeling methods, which were successful in similar settings, but resulted in low agreement scores in our setup.
We highlight some of the issues we believe are causing the disagreements. The main one being different construals \cite{trott2020construing} of the utterances by different people --- as well as the difficulty to consider a different one, once fixating on a specific construal --- that led to different answers.
We connect our findings to previous work that observed some inherent disagreement in human judgments in popular datasets, such as SNLI and MNLI \cite{pavlick2019inherent}.
Although this issue is less prominent in these datasets (which is manifested as higher agreement scores), we notice that when tackling a {\em specific} phenomenon, e.g. involving implicit elements, these issues may arise. 

We also argue that the {\em lack} of detailed definitions in the commonly used NLI tasks may lead to poor performance on small buckets of language-specific phenomena. This drop might be lost in large-scale datasets, but may have critical effects when modeling and studying specific phenomena.
As a community, we claim, we should seek to identify those buckets and further investigate them, using more profound approaches for data collection, with clear and grounded definitions. We hope that our attempted trial in data collection will allow others to learn from our failure.

\section*{Acknowledgments}
We would like to thank Adam Poliak and Abhilasha Ravichander for providing valuable feedback on this paper.
Moreover, we would like to thank the reviewers, as well as the workshop organizers for their constructive reviews.
Yanai Elazar is grateful to be partially supported by the PBC fellowship for outstanding Phd candidates in Data Science.
This project has received funding from the Europoean Research Council (ERC) under the Europoean Union's Horizon 2020 research and innovation programme, grant agreement No.\ 802774 (iEXTRACT) and grant agreement No.\ 677362 (NLPRO).

\bibliography{emnlp2020}
\bibliographystyle{acl_natbib}

\clearpage

\appendix

\section{Linguistic Background}
\label{sec:app_ling}

Complement coercion has been studied in linguistics from many theoretical viewpoints.
 Lexical semantic accounts (such as \citealt{Pustejovsky1991TheGL,pustejovsky1995gl} and others) and Construction Grammar accounts (e.g. \citealt{Goldberg95}) ``attempt to formalize what semantic features of a lexical item have been changed to conform to those of the construction” \cite{Yoon2012ConstructionsSC}.
One of the main approaches  is the Type-Shifting analysis \cite{Pustejovsky1991TheGL,pustejovsky1995gl,Jackendoff1996TheAO,Jackendoff2002FoundationsOL},
``which asserts that complement coercion involves a type-shifting operation that coerces the entity-denoting complement to an event”\cite{yao-ying_2017}. Another approach (\citealt{Almeida2008CoercionWL} and others) ``claims that complement coercion involves a hidden VP structure with an empty verb head, which is saturated by pragmatical inference in context” \cite{yao-ying_2017}. 
Cognitive linguistics accounts (such as \citealt{Kovecses1998MetonymyDA}) exploit metonymy as the mechanism behind coercion constructions \cite{Yoon2012ConstructionsSC}.
Complement coercion has been also extensively investigated in the framework of neurolinguistic research (for example, \citealt{Kuperberg2010ElectrophysiologicalCO}) and psycholinguistic studies (e.g., \citealt{McElree2006ATC}). The latter often show that ``coercion sentences elicit
increased processing times" \cite{Husband2011UsingCC} compared with non-coercion sentences. Such theories as the Type-Shifting Hypothesis mentioned above and the Structured-Individual Hypothesis \cite{Piango2016ReanalyzingTC} suggest different explanations for this associated processing cost \cite{yao-ying_2017}.

\section{Complement Coercion: Counter Examples}
\label{sec:app_neg}

Here we provide some additional examples of constructions that are similar to the ones in Examples \ref{ex1},\ref{ex2} (the verb `start' is followed by a non-event-denoting complement) but do \emph{not} function as complement coercion constructions.
Consider the following sentences:
\begin{enumerate}
\setcounter{enumi}{6}
\item \label{ex1-app} I \textbf{started} a new company.
\item \label{ex2-app} His name \textbf{started}  the list.
\item \label{ex3-app} Her wedding dress  \textbf{started}  a new tradition among brides.
\end{enumerate}
In example \ref{ex1-app} the verb `start' is used as an entity-selecting (and not event-selecting) verb, a synonym of `found', `establish', so that there is no type clash.

In  example \ref{ex2-app} the verb `start' is used in its `non-eventive' \cite{Zarcone}  or `stative' \cite{Pinango} sense ('constitute the initial part of something'). When used this way, the verb `start' does not exclusively select for eventive complements, so, again, there is no type clash. Also, some authors \cite{godard-jayez-1993-towards,yao-ying_2017,pustejovsky-bouillon-1994-proper} argue that in coercion constructions the subject should be an ``intentional controller of the event'' \cite{godard-jayez-1993-towards}. In example \ref{ex3-app} this condition does not hold, therefore there is no coercion.

\section{Explicit Modeling}
\label{sec:app:explicit}

\begin{figure}[t!]
\centering

\includegraphics[width=0.95\columnwidth]{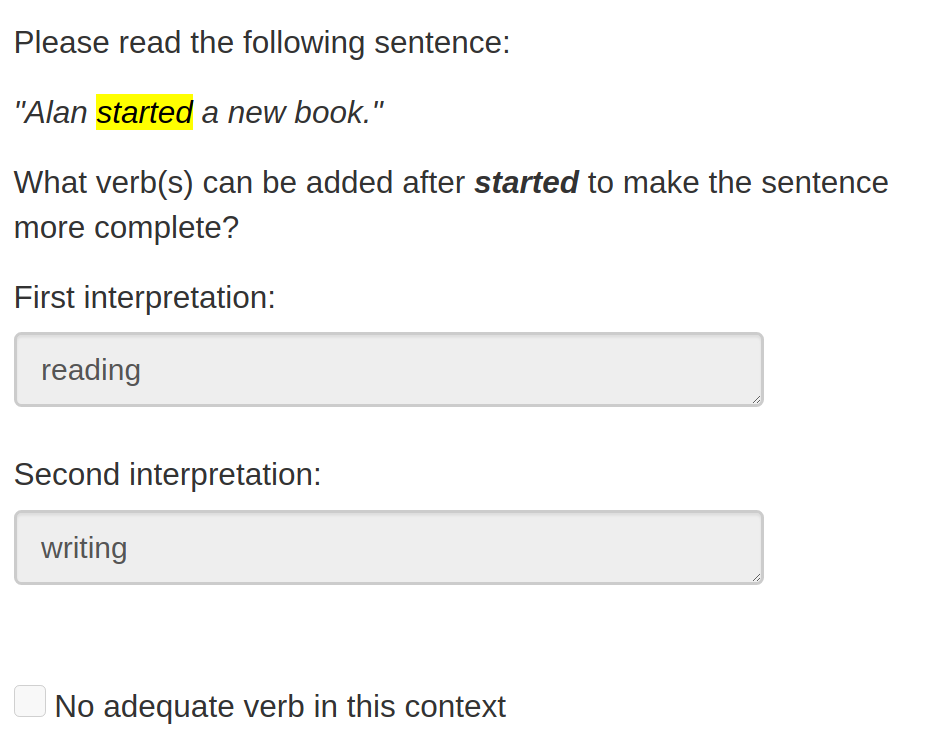}

\caption{A screenshot of the explicit task presented to the annotators.}
\label{fig:explicit}
\end{figure}

In the \textit{Explicit Completion} approach, the goal is to add the implicit argument of the coercion construction, if such completion exists. For instance, in the sentence ``I started \emp a new book'', possible completions are `reading' and `writing', and in Example \ref{ex1-app} no completion fits.
Concretely, given a sentence with a complement coercion verb candidate, the task is to complete it with a set of possible verbs that describe the covert event. As not all candidates function as parts of complement coercion constructions, annotators can mark that no additional verb is adequate in the context.
In cases where there is more than one semantically plausible answer (e.g. Ex. \ref{ex1}), we ask annotators to provide two completion sets, each consisting of a group of semantic equivalent verbs, which correspond to different possible understandings of the text.
A screenshot of the task presented to the turkers is shown in Figure \ref{fig:explicit}.

This approach to task definition is reminiscent of those used for other missing elements phenomena, such as Verb Phrase Ellipsis \cite{bos2011ellipsis}, Numeric Fused-Heads \cite{nfh}, Bridging \cite{BASHI_bridging,Hou_thesis_bridging} and Sluicing \cite{hansen2020sluicing}.
However, in contrast to these tasks, where the answers can usually be found in the context,\footnote{Although not always. Some of the answers in the NFH work by \citet{nfh} are also open-ended, but those are relatively rare. Furthermore, the answers in sluicing are sometimes a modification of the text.} the answers in our case are more open-ended (although still bounded by some restrictions \cite{godard-jayez-1993-towards,pustejovsky-bouillon-1994-proper}). This makes this task more challenging for annotation.

\paragraph{Corpus Candidates}
In the \textit{explicit completion} setting, we look for natural sentences that contain one of the following anchor verbs: `start', `begin', `continue' and `finish', - immediately followed by a direct object without any dependent verb in between. 

\paragraph{Annotation Procedure}
We use the same restrictions from the previous procedure and create a new validation test, tailored for the new task. We pay 4 cents per Hit.

\paragraph{Results}
We collect annotations for 200 sentences, with two annotations per sentence. We compute the Fleiss Kappa \cite{fleiss_kappa} after a relaxation of the annotations into two labels: added a complement or not.
Similarly to the previous modeling, the agreement score is $k=0.18$, which is considered to be low. Consider the following examples:

\begin{enumerate}
\setcounter{enumi}{8}
    \item ``In 2011, Old Navy \textbf{began} \emp a second rebranding to emphasize a family-oriented environment, known as Project ONE.'', --- \textit{$\{advertising,promoting,endorsing\}, \phi$}
    \item ``After he had \textbf{finished} \emp his studies Sadra began to explore unorthodox doctrines and as a result was both condemned and excommunicated by some Shi'i `ulamā'.'', --- \textit{$\{pursuing, doing\}, \phi$}
\end{enumerate}

According to the definition of complement coercion, these examples do not require a complement. However, as can be seen from these examples, the proposed complements do contribute to an easier understanding of the sentence. We note that this concept of `missing' is hard to explain and can be also subjective. 
Another obstacle is that strict adherence to the linguistic definition does not always contribute to potential usefulness of the task for downstream applications. 
For this phenomenon, we did not follow the strict linguistic definition and used a more relaxed one.
Additional examples along with their annotations are provided in Table \ref{tbl:explicit}.

\section{NLI Framing: Additional Material}
\label{sec:nli-additional}

\begin{figure}[t!]
\centering

\includegraphics[width=0.95\columnwidth]{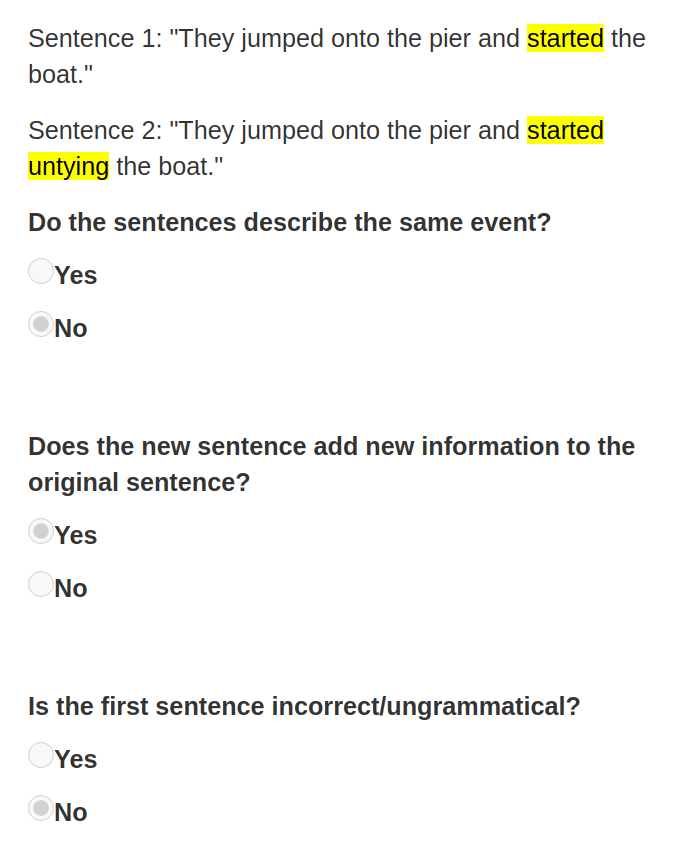}

\caption{Screenshot of the interface shown to the turkers for collecting labels. This setup follows the instructions used for labeling NLI data in \citet{glockner-vered-nli}.}
\label{fig:nli}
\end{figure}

We provide a screenshot of the NLI interface shown to the turkers in Figure \ref{fig:nli}.

\paragraph{NLI Data}

We provide additional examples for the original and the modified sentences (hypotheses and premises accordingly) used in the NLI framing (\S \ref{sec:nli-cc}), along with the three obtained labels, in Table \ref{tbl:more-nli}.

\begin{table*}[ht!]
\centering
\resizebox{\textwidth}{!}{%

\begin{tabular}{lll}
\toprule
Premise & Hypothesis & Annotations \\
\midrule

that gives us something to work with if he starts trouble.  & that gives us something to work with if he starts making trouble. 
& \entlabel{} \entlabel{} \entlabel{} \\

I do hope you will continue mrs. cox's incredible hospitality. & I do hope you will continue to enjoy mrs. cox's incredible hospitality. & \conlabel{} \conlabel{} \conlabel{} \\

he asked me as he continued a tune.
& he asked me as he continued to strum a tune.
& \neulabel{} \neulabel{} \conlabel{} \\

how would she continue questions like this? & how would she continue to answer questions like this? & \conlabel{} \conlabel{} \conlabel{} \\

he finished a sip of coffee and replied, not surprised. & he finished taking a sip of coffee and replied, not surprised. & \entlabel{} \entlabel{} \neulabel{} \\

it was pike's idea to start these games. & it was pike's idea to start playing these games. & \entlabel{} \neulabel{} \conlabel{} \\

I started deep breaths and tried to cleanse my mind. & I started taking deep breaths and tried to cleanse my mind. & \entlabel{} \entlabel{} \neulabel{} \\

I would like to finish this movie sometime in this year! & I would like to finish watching this movie sometime in this year! & \entlabel{} \entlabel{} \conlabel{} \\

\midrule

\bottomrule
\end{tabular}
}
\caption{Examples for NLI pairs with a complement coercion structure.
The \entlabel{}, \neulabel{} and \conlabel{} labels refers to the entail, neutral and contradict accordingly.}
\label{tbl:more-nli}

\end{table*}

\begin{table*}[t!]
\centering
\resizebox{\textwidth}{!}{%

\begin{tabular}{ll}
\toprule
 Text &  Annotations \\
\midrule

... it will likely travel in a parabola, \textit{continuing} its stabilizing spin, ... & $\phi, \phi$ \\

Afterwards, they decide to \textit{continue} the pub crawl to avoid attracting suspicion. & $\{doing\}, \{doing\}$ \\

I was surprised he did not \textit{continue} his openness at the RFPERM. & $\{embue\}, \{showing, displaying, ...\}$ \\

In 1994, he joined Motilal Oswal to \textit{start} their institutional desk before moving to UBS in 1996. & $\{employ\}_1,\{work\}_2, \{working\}$ \\

In 1943 she \textit{started} a career as an actress with the stage name Sheila Scott a name ... & $\phi, \{pursuing\}$ \\

..., giving him the opportunity to \textit{continue} the work left by his predecessors as well as ... & $\phi, \{researching, studying\}$ \\

In the Middle Ages it was a battle cry , which was used to \textit{start} a Feud or a Combat reenactment. & $\phi, \{fighting\}$ \\

In addition, deductions are taken if the man \textit{finishes} the element on two feet ... & $\phi, \{competing\}$ \\

\bottomrule
\end{tabular}
}
\caption{Examples for the Explicit modeling. $\phi$ denotes the empty set, meaning no event is implied. When a subscript is present it denotes the different interpretation of the sentence, by the same annotator.}
\label{tbl:explicit}

\end{table*}

\end{document}